%% file: main.tex
\documentclass[conference]{IEEEtran}
\IEEEoverridecommandlockouts
\usepackage[T1]{fontenc}
\usepackage[utf8]{inputenc}
\usepackage{lmodern,graphicx,amsmath,amssymb,booktabs,array,tabularx}
\usepackage[numbers,sort&compress]{natbib}
\usepackage{url,microtype,stfloats,placeins}
\usepackage{xcolor}
\definecolor{linkblue}{HTML}{24557A}
\usepackage[colorlinks=true,linkcolor=linkblue,citecolor=linkblue,urlcolor=linkblue]{hyperref}
\hypersetup{pdftitle={Designing Informative Actions: Action Construction for Long-Horizon Research Agents},pdfauthor={Alkut Arxidin, Shaoning Zeng, Jianhang Zhou, Jianping Gou, Xinzhong Zhu}}
\newcommand{\IG}{\operatorname{IG}}
\newcommand{\E}{\mathbb{E}}
\newcommand{\ind}{\mathbf{1}}
\newcommand{\score}{Q_B}
 \newcommand{\abbr}[1]{{\upshape #1}}
\input{generated/numbers}

\begin{document}
\title{Designing Informative Actions: Action Construction for Long-Horizon Research Agents}
\author{\IEEEauthorblockN{Alkut Arxidin, Shaoning Zeng\textsuperscript{*}, Jianhang Zhou, Jianping Gou, Xinzhong Zhu}
\IEEEauthorblockA{\textsuperscript{*}Corresponding author}}
\maketitle
\pagestyle{plain}
\thispagestyle{plain}
\raggedbottom

\begin{abstract}
Large language model agents support automated scientific research by connecting literature analysis and planning with experimentation. Evaluating which action an agent selects does not establish which experiment its instruction specifies, because omitted interventions or replication counts leave consequential choices to the executor. We introduce a reading-set framework for evaluating action construction under a common cost rule. Across 30 synthetic states, complete-design requests increase affordable efficiency by \RelativeGainA\% and \RelativeGainB\% relative to the action-name baseline in the two model families, respectively. An author-constructed research-action evaluation dataset provides 60 scenarios for complementary judgment evaluation. On this dataset, the scenario-level information-per-cost measure has a mean within-scenario correlation of \RhoHuman{} with judgments from 18 human participants. The framework makes design specification and execution ambiguity measurable, providing a reproducible basis for assessing how research agents translate scientific intent into informative, budget-feasible experiments.
\end{abstract}


\section{Introduction}
Large language model agents are expanding the scope of automated scientific research. Surveys chart their growing role in literature analysis, hypothesis generation, and experimentation \cite{ref14,ref15}. Systems including the AI Scientist, AI co-scientist, Agent Laboratory, and Kosmos connect these capabilities into integrated research workflows \cite{ref10,ref11,ref12,ref13}.

Research-agent benchmarks assess experimental performance, scientific rediscovery, and innovation, while workflow and process evaluations examine intermediate decisions and verification targets \cite{ref16,ref18,ref22,ref19,ref29,ref30}. Bayesian experimental design values specified experiments by expected information gain, and cost-aware learning incorporates acquisition expense into resource allocation \cite{ref1,ref2,ref3,ref7}. Tool-using scientific agents connect language-model plans to executable operations, while studies of prompt sensitivity and structured outputs examine the interfaces through which those plans are expressed \cite{ref8,ref9,ref25,ref26,ref27,ref28,ref41}. Yet selecting an appropriate research action does not establish which experiment its instruction specifies. Language-grounding and clarification studies show why execution can require choices beyond the instruction itself \cite{ahn2022saycan,ren2023knowno}. Experimental utility depends on the specified design and its outcome model \cite{ref1,ref2,ref3}, so different executable interpretations of the same instruction can have different information values. The unresolved question is how to evaluate the information value and affordability of an instruction when it leaves experimental choices to the executor.

The central idea is therefore to evaluate the designs admitted by the instruction, making this dependence visible. We propose a reading-set framework that scores those designs under a common cost rule with per-design budget gating. Across 30 synthetic states, complete-design requests increase affordable efficiency by \RelativeGainA\% and \RelativeGainB\% relative to action-name requests in the two model families. Complementary evidence comes from an author-constructed dataset of 60 scenarios and 240 candidate actions with traceable source records. On this dataset, the scenario-level information-per-cost measure shows positive agreement with judgments from a six-model panel and 18 human participants.

\section{Related Work}
\label{sec:related}

\subsection{Research Agents and Scientific Evaluation}

Research-agent systems span hypothesis generation, literature analysis, implementation, and experimentation \cite{ref10,ref11,ref12,ref13}. General agent benchmarks and evaluation surveys organise assessment around interactive task performance and evaluation objectives \cite{ref31,ref32}. Benchmarks make these capabilities measurable through tasks involving experimental performance, information organisation, rediscovery, and innovation \cite{ref16,ref17,ref18,ref22}. At the workflow level, systematizations map research workflows to evaluation and verification targets \cite{ref19}. At a finer scale, trace datasets and process reward models examine intermediate behaviour and reasoning \cite{ref29,ref30}. In controlled agent environments, exploration and exploitation errors can also be measured separately from observed actions \cite{ref21}. These approaches establish complementary views of research performance, but task success and step-level correctness do not by themselves resolve the experimental choices left unspecified in an action. A scientifically relevant instruction can still admit several executable designs with different information values and costs.

\subsection{Information- and Cost-Aware Experimental Design}

Bayesian experimental design uses expected information gain (IG) to identify experiments that reduce uncertainty about competing hypotheses \cite{ref1,ref2,ref3}. Bayesian optimisation supports efficient search over expensive evaluations \cite{ref4,ref5,ref6}, while cost-aware active learning incorporates the expense of acquiring additional evidence \cite{ref7}. Although these objectives provide a principled basis for allocating experimental resources, applying them to a natural-language action still requires identifying the executable design or designs it denotes. When interventions or replication counts are omitted, a single utility value depends on how those omissions are completed. The reading-set framework addresses this dependence by making the admitted designs explicit before aggregating their budget-gated information-per-cost contributions.

\subsection{From Natural-Language Instructions to Executable Actions}

Autonomous chemistry systems connect language-model planning with scientific tools and laboratory operations, demonstrating how tool interfaces can carry research intentions into executable workflows \cite{ref8,ref9}. In robotic planning, SayCan grounds action selection in skill feasibility \cite{ahn2022saycan}, while KnowNo uses calibrated candidate sets to determine when an ambiguous instruction requires clarification \cite{ren2023knowno}. Procedural skill libraries encode reusable research workflows as instructions and supporting resources \cite{ref20}. Research on prompting shows that ordering, formatting, and contextual features can influence model responses \cite{ref25,ref26,ref27,ref28}, and work on schema evolution examines the reliability of structured outputs \cite{ref41}. However, the reliability of an action interface does not by itself establish whether the proposed experiment is fully specified, informative, and affordable. A parsable instruction can still omit interventions or replication counts, leaving the executor to determine the design that is ultimately evaluated.

\section{Method}
\label{sec:method}
\begin{figure*}[t]
\centering\includegraphics[width=\textwidth]{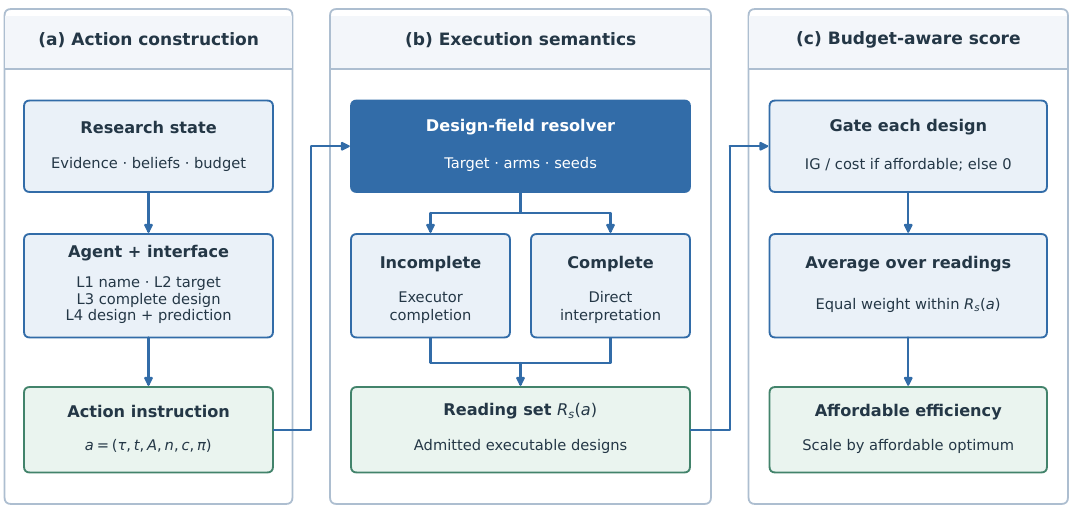}
\caption{Action construction from research intent to a budget-aware design score. (a) The agent receives a research state and an action interface. (b) Parsed fields determine a reading set through executor completion or direct interpretation of a valid complete design. (c) Each design contributes IG/cost if affordable and zero otherwise, before equal-weight averaging and normalisation by the best affordable information efficiency.}
\label{fig:method}
\end{figure*}

The framework evaluates the experimental choices communicated through an action interface. Given a research state and a model response, the parser extracts the design fields and the resolver maps them to a set of executable designs. We then apply a budget gate to each design, average its information-per-cost contribution, and normalise the result by the best affordable information efficiency. This flow connects action specification, executable interpretation, and scoring, as shown in Fig.~\ref{fig:method}. The common cost rule and aggregation are defined in Eqs.~\eqref{eq:cost}--\eqref{eq:score}.

\subsection{States, Actions, and Executable Designs}
A research state $s$ contains a scientific question, a method description, partial results, candidate mechanisms, a belief distribution $p(h\mid s)$ over mechanisms, and a budget $B$. An executable design is $d=(A,n)$, where $A$ is a nonempty set of ablation interventions and $n$ is the number of seeds per condition. The unablated baseline is always run. The synthetic benchmark uses six interventions and $n\in\{2,\ldots,12\}$, giving $63\times11=693$ candidate designs per state.

We represent an action as $a=(\tau,t,A,n,c,\pi)$, comprising action type $\tau$, target $t$, intervention set $A$, replication count $n$, success criterion $c$, and optional prediction $\pi$. The four interfaces specify different design information. The \textbf{action-name interface (\abbr{L1})} requests an action name and a reason, but its parser extracts only the action name. The \textbf{target interface (\abbr{L2})} requests a target and its justification, leaving interventions and replication to the executor. By contrast, the \textbf{complete-design interface (\abbr{L3})} requests the target, intervention set, replication count, and success criterion. The \textbf{design-with-prediction interface (\abbr{L4})} adds a prediction to these complete-design fields. Target, arms, and seed count determine the reading set, whereas success criteria and predictions remain recorded text. Thus, two outputs with the same scored design parameters receive the same design score even if their explanations differ. Appendix~\ref{app:reproduce} specifies the exact fields and parser interpretations.

\subsection{Reading Sets and Executor Sensitivity}
\label{sec:readings}
Let $R_s(a)$ denote the designs produced by the fixed executor for action $a$ in state $s$. The default evaluator weights its members uniformly. For an action without a stated target, the component-scan reading set contains four designs. These use one, two, or four interventions at two seeds, or one intervention at four seeds. With a target but no design, a fixed target-focus rule selects interventions and considers two replication levels. Complete arms and seed fields identify one design. Appendix~\ref{app:reproduce} records the resolver details. Appendix~\ref{app:example} illustrates how individual responses are resolved and scored.

Two additional executor modes examine completion sensitivity. The conservative mode keeps the cheaper half of the candidate set, whereas the permissive mode keeps the more expensive half. Because neither mode changes a singleton set, a fully specified design remains the same in all modes. We also record the ambiguity index $g(a)=1-1/|R_s(a)|$. Its value is zero for a singleton and increases with the number of admitted readings. This index measures set cardinality, rather than the fraction of experimental parameters invented by an executor.

\subsection{Information, Cost, and the Primary Score}
\label{sec:score}
We use the information-theoretic utility of Bayesian experimental design \cite{ref1,ref2,ref3}. A design is informative when its possible outcomes reduce uncertainty about competing mechanisms. Accordingly, its information value is the expected reduction in mechanism entropy, defined as follows.
\begin{equation}
\IG_s(d)=H(p(h\mid s))-\E_{o\sim p(o\mid s,d)}H(p(h\mid s,d,o)).
\end{equation}
Entropy uses natural logarithms, giving information in nats. The observation model is a normalised Gaussian kernel on the discrete grid $\mathcal{Y}=\{0,0.25,0.5,0.75,1\}$. For intervention $j$, the probability of an observation is given by the following normalised kernel.
\begin{equation}
p(y_j\mid h,d)=\frac{\exp[-(y_j-\mu_{hj})^2/(2\sigma_n^2)]}{\sum_{z\in\mathcal{Y}}\exp[-(z-\mu_{hj})^2/(2\sigma_n^2)]}.
\end{equation}
Here $\sigma_n=0.55\sqrt{5/n}$ and $\mu_{hj}\in\{0,1\}$ encodes whether the intervention removes the hypothesised source of the improvement. Outcomes are conditionally independent given $h$. The hypothesis-independent baseline factor cancels from the posterior. The evaluator sums over the joint outcome grid in log space, omitting outcome probabilities at or below $10^{-10}$. This is a finite observation model with numerical expected information gain.

The cost model accounts for experiment setup, interventions, and replication. The prompt exposes the same rule used by the evaluator, so the model can assess affordability before answering. We define the design cost as follows.
\begin{equation}
\label{eq:cost}
C(d)=1+0.6|A|+0.35n.
\end{equation}
The intervention count $|A|$ excludes the baseline. The fixed setup term includes the baseline convention. This benchmark cost is a dimensionless resource model, not measured GPU time. For example, six interventions at nine seeds cost $7.75$ units and fit a budget of $8$.

Budget feasibility is checked for each reading before aggregation. An affordable reading contributes its information per cost, while an unaffordable one contributes zero. The action utility averages these contributions and the primary score normalises this average by the best affordable design.
\begin{align}
U_B(a\mid s)&=\frac{1}{|R_s(a)|}\sum_{d\in R_s(a)}\ind[C(d)\le B]\frac{\IG_s(d)}{C(d)},\\
\score(a\mid s)&=\frac{U_B(a\mid s)}{\displaystyle\max_{d\in\mathcal{D}:C(d)\le B}\IG_s(d)/C(d)}.
\label{eq:score}
\end{align}
The primary outcome, \emph{affordable efficiency share}, is $100\score$. An unaffordable reading contributes zero while remaining in the reading-set denominator. Empty or invalid final answers score zero. The reference searches all 693 designs, including every allowed integer seed count. Because every nonzero term in the numerator is bounded by the affordable reference, $0\le\score\le1$. Applying the gate to individual readings also keeps an affordable average cost from masking an unaffordable member of the set.

\section{Experiments}
\label{sec:experiments}

\subsection{Datasets and Experimental Setup}
\label{sec:synthetic-design}
We use two datasets to study complementary aspects of action construction. For controlled comparisons, we generate 30 synthetic research states with a fixed random seed of 11. Each state describes a Transformer with six candidate components, partial performance changes on three or four simulated datasets, candidate mechanism signatures, a stated prior, and a budget of $4$, $6$, or $8$ units. The generator varies which datasets improve and by how much, then assigns prior mass according to each mechanism signature's overlap with those improvements. These simulated observations define the decision context. All 693 intervention--replication designs are scored under the finite observation model in Section~\ref{sec:score}.

For judgment evaluation, we use our \emph{author-constructed research-action evaluation dataset}. It contains 60 scenarios and 240 candidate actions, with 20 scenarios each in machine learning, systems, and AI for science. The construction draws on observations and component comparisons in 29 published source papers. Source records identify papers by DOI or arXiv identifier and locate the relevant observations and component comparisons by section, figure, or table. We adapt these materials into a research question, baseline, partial observation, six candidate mechanisms and corresponding interventions. Construction checks cover source locatability, mechanism--intervention correspondence, and separation of visible prompts from private source conclusions. Each scenario is paired with four constructed actions that vary target, intervention coverage, replication, and specification completeness, giving 240 candidates. Source observations are kept separate from project-defined priors, abstract budgets, and simulator mappings. Proposed follow-up interventions carry no newly measured outcomes. Scenarios can share a source paper, so the dataset contributes 60 traceable decision contexts rather than 60 independent source studies. Appendix~\ref{app:external} details construction and provenance.

The synthetic benchmark compares the four interface levels \abbr{L1}--\abbr{L4}. DeepSeek (family A) is accessed through the deployment deepseek-flash, with the reasoning-effort parameter set to low or max. Qwen3.8-27B (family B) is evaluated with the parameter set to low or omitted. The label na denotes parameter omission, not a specified reasoning-effort level; the provider determines the effective setting. Every request supplies the budget, Eq.~\eqref{eq:cost}, intervention identifiers, and allowed seed range, and uses a 65,536-token ceiling. Three responses per state and condition give 1,440 main outputs. Concise and expanded interface controls add 360 outputs, and a design-format control adds 180. A separate complete-design prompting comparison contributes 540 final outputs and 180 first-turn drafts. All 2,520 final items in these analyses are scored, including eight invalid answers assigned zero.

Affordable efficiency share is the primary synthetic outcome, while parsing and feasibility rates describe executability. Alternative executor modes and deterministic design policies examine completion dependence and design quality. Because requested content and parsed fields change together, the main estimand is the information delivered by the complete request--parser interface. We average the three responses within each state, use 5,000 state-bootstrap resamples for 95\% confidence intervals (\abbr{CIs}), and apply Holm correction to 12 planned paired contrasts. Appendix~\ref{app:reproduce} provides collection and parsing details, quality counts, and the full testing and resampling specifications.

The constructed scenarios evaluate agreement with the separate scenario-level IG/cost measure. Six open-weight models rank the four candidates under four cyclic position assignments, producing 1,440 ranking responses. Position balancing addresses order sensitivity in preference judgments \cite{ref23,ref24,ref28}. Eighteen participants who self-report graduate research experience each assess eight scenarios through all six action pairs, yielding 864 judgments and 216 attention checks. We report pooled correlations across the 240 actions and mean within-scenario correlations across sets of four actions, with scenario-bootstrap intervals. Thus, this study tests agreement about proposed research actions, whereas the synthetic study tests affordable efficiency under its explicit simulation model.

\subsection{Interface and Baseline Comparisons}

\label{sec:mainresults}
\label{sec:synthetic-results}\label{sec:baselines}
\begin{figure*}[!t]
\centering\includegraphics[width=\textwidth]{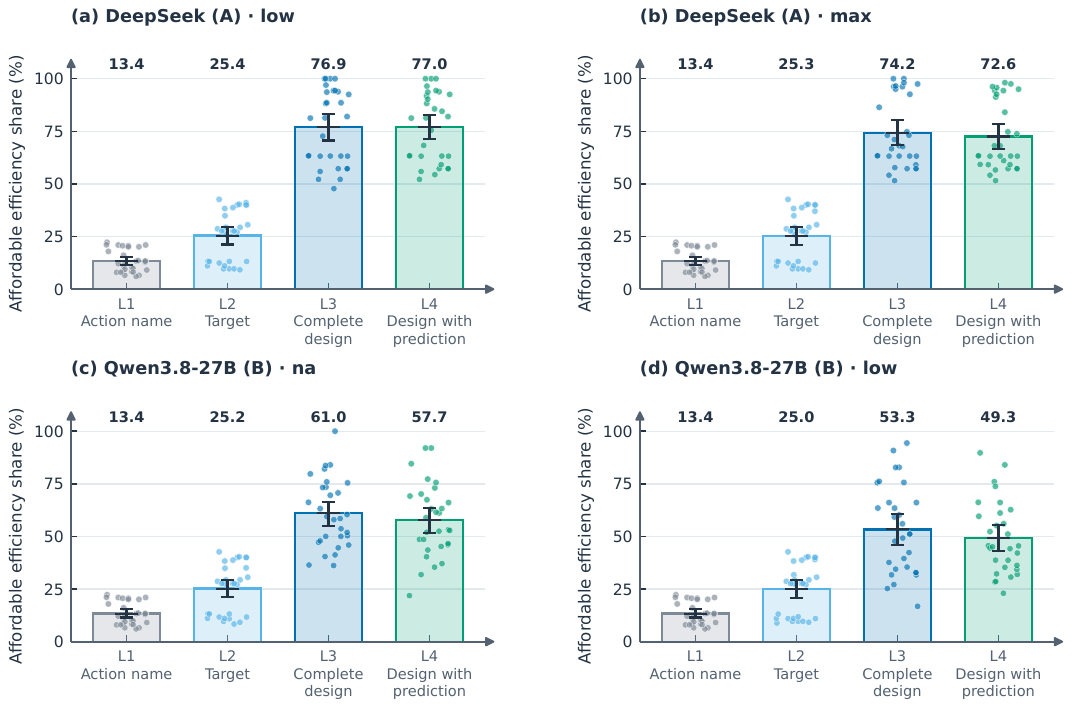}
\caption{Complete-design interfaces increase affordable efficiency across both model families. Bars and top labels show means; whiskers show 95\% state-bootstrap confidence intervals. Small points show all 30 state means, each averaging three responses; horizontal jitter separates points. Panels share a 0--100\% scale. All 90 requests per condition are included, with invalid answers scored as zero. The label na means that the reasoning-effort parameter is omitted.}
\label{fig:main}
\end{figure*}

We first test whether explicitly requesting design parameters improves the affordable experiment delivered to the executor. As shown in Table~\ref{tab:main} and Fig.~\ref{fig:main}, complete-design requests substantially improve affordable information efficiency in both model families. On DeepSeek/low, \abbr{L1} scores \ScoreLone\%, \abbr{L2} 25.4\%, and \abbr{L3} \ScoreAlowLthree\%. \abbr{L4} scores 77.0\%. The corresponding \abbr{L3} scores are \ScoreAmaxLthree\% for DeepSeek/max, \ScoreBnaLthree\% for Qwen3.8-27B/na, and \ScoreBlowLthree\% for Qwen3.8-27B/low. The increase appears in every tested setting, connecting explicit design specification to higher expected information per available resource.

\begin{table*}[t]
\centering\small
\caption{Primary affordable efficiency share (\%). Each cell contains 30 states and three responses per state. All failed outputs remain in the denominator. The same budget gate, cost model, and complete affordable reference apply to all conditions. The label na means that the reasoning-effort parameter is omitted.}
\label{tab:main}
\input{generated/main}
\end{table*}

Averaging the two settings within each family, affordable efficiency rises from \FamilyBaseA\% to \FamilyCompleteA\% for DeepSeek and from \FamilyBaseB\% to \FamilyCompleteB\% for Qwen3.8-27B. These correspond to relative increases of \textbf{\RelativeGainA\%} and \textbf{\RelativeGainB\%}, calculated as $100(\bar Q_{\mathrm{L3}}-\bar Q_{\mathrm{L1}})/\bar Q_{\mathrm{L1}}$. Paired tests on the underlying state-level score differences yield $p_{\mathrm{Holm}}=7.42\times10^{-17}$ for DeepSeek and $7.66\times10^{-15}$ for Qwen3.8-27B. Appendix~\ref{app:reproduce} reports all planned contrasts in the original 0--1 score units.

The efficiency gains occur under the stated budgets. Among the \DesignTotal{} main \abbr{L3}/\abbr{L4} requests, \DesignFeasible{} yield a parsed design within budget (\FeasiblePct\%). All 360 DeepSeek complete-design requests are feasible and parseable, compared with 355 of 360 for Qwen3.8-27B. Because the primary score assigns zero to an unaffordable reading or an invalid output, the reported advantage already includes those outcomes.

The source of this separation is visible in the executor inputs. At the action-name interface (\abbr{L1}), the four settings are identical state by state: their parsed action names lead to the same executor completions. This equality characterises the interface's expressed information. Once models supply arms and seeds, their scores reflect distinct constructed experiments. The evaluation therefore reveals where a planning interface begins to expose design choices to measurement.

We next compare complete-design prompting strategies on the same 30 synthetic states. Think-then-act is a single-decision adaptation inspired by ReAct \cite{yao2023react}, the planning baseline adapts Plan-and-Solve \cite{wang2023plan}, and one-round Self-Refine adds a critique and revision to an independently generated draft \cite{madaan2023selfrefine}. All three request a target, interventions, replication count, and success criterion and use the same strict complete-design parser as \abbr{L3}. Each strategy contributes 90 final responses per model setting, with failures retained. The new baseline responses and the original \abbr{L1}--\abbr{L4} responses match state identities, model request settings, cost rules, and scoring, while their collection batches differ.

On DeepSeek with maximum reasoning effort, \abbr{L3} attains 74.25\% affordable efficiency, compared with 70.97\% for think-then-act, 70.98\% for planning, and 71.88\% for self-refinement. The design-with-prediction interface (\abbr{L4}) scores 72.56\%, whereas \abbr{L1} and \abbr{L2} score 13.36\% and 25.29\%, respectively. The complete-design interface has the highest observed mean in this setting, as summarised in Table~\ref{tab:prompts}. The full two-model comparison, including model-dependent strategy rankings, is reported in Appendix~\ref{app:complete-baselines}. This comparison separates the value of providing an executable design from the choice of prompting strategy used to produce it.

\begin{table}[t]
\centering\footnotesize
\caption{Complete-design prompting strategies and the four study interfaces on DeepSeek with maximum reasoning effort. Efficiency is affordable efficiency share (\%). Each method includes all 90 final responses across 30 states. Bold method names identify our four study interfaces. Bold numerical values mark the highest mean.}
\label{tab:prompts}\input{generated/prompt_baselines}
\end{table}

\subsection{Judgment Agreement on the Constructed Dataset}

\label{sec:external}

\begin{figure}[t]
\centering\includegraphics[width=\columnwidth]{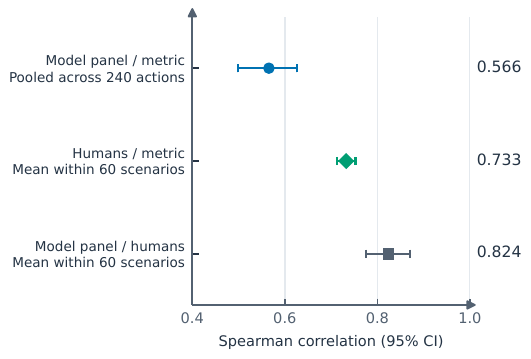}
\caption{Agreement of the scenario-level IG/cost measure with research judgments. Markers indicate Spearman correlations and horizontal bars show 95\% scenario-bootstrap confidence intervals. Labels distinguish pooled from mean within-scenario statistics. The estimates describe judgment agreement, not realised experimental outcomes.}
\label{fig:external}
\end{figure}

We examine whether the scenario-level IG/cost measure ranks the constructed candidate actions consistently with model-panel and human judgments. Model ranks are first averaged over the four positions and then across the six models. Similarly, human ranks are averaged within each scenario before correlation. The model panel has a pooled Spearman correlation of \RhoS{} with the external metric. Human--metric agreement averages \RhoHuman{} within scenarios. The model panel and humans have a mean within-scenario correlation of \RhoSH{}. These positive correlations support agreement between the metric and research judgments on this constructed dataset. The three agreement estimates and their uncertainty are shown in Fig.~\ref{fig:external}. Because the pooled and scenario-averaged statistics use different aggregation units, their magnitudes should not be interpreted as directly comparable effect sizes. Appendix~\ref{app:external} reports the full estimates and judgment protocol.

Table~\ref{tab:panel-models} reports agreement separately for each panel model. All six models show positive pooled correlations with the metric, ranging from 0.244 to 0.588. At the scenario level, their mean correlations range from 0.217 to 0.594. Agreement therefore appears across the panel's constituent models, with variation in strength. The panel's pooled correlation is 0.566 and its mean within-scenario correlation is 0.545. These panel values are computed after averaging model ranks, rather than by averaging the six correlation coefficients.

\begin{table*}[t]
\centering\small
\caption{Model-panel agreement with the scenario-level IG/cost measure, resolved by model. Each model contributes 240 valid ranking responses: four position assignments in each of 60 scenarios. Pooled correlations use all 240 action ranks after position averaging. Within-scenario correlations compare the four actions and are averaged over the defined scenarios. Yi has one constant position-averaged rank vector, so its within-scenario mean uses 59 scenarios. All of its responses remain in the pooled and panel calculations. The panel row averages ranks across all six models before correlation.}
\label{tab:panel-models}\input{generated/panel_models}
\end{table*}

The name-only and minimally specified proposals also reveal an ordering difference. Human and panel raters prefer the minimally specified text, while the metric can favour the name-only action after completing its design. Both actions have nonempty reading sets and positive information values. This difference depends on the object being evaluated, as discussed in Section~\ref{sec:discussion}.

\subsection{Ablation and Interface Controls}

\label{sec:ablation}
The final experiment examines which interface elements accompany the construction gain. We compare \abbr{L3} with \abbr{L4} to assess adding a prediction field, and use concise and expanded requests to test alternative presentations. These are interface ablations and controls. The score reads design parameters, so the contrasts do not isolate an underlying scientific-reasoning capability.

\subsubsection{Concise and Expanded Requests}\label{sec:concise}
\begin{figure*}[t]
\centering\includegraphics[width=\textwidth]{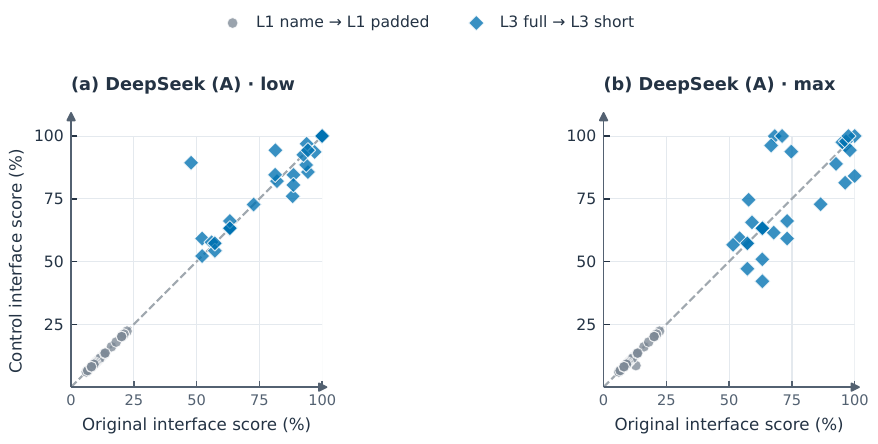}
\caption{Original and control interface scores, paired by state. Each point averages three responses per axis for \abbr{L1} name versus \abbr{L1} padded, or \abbr{L3} full versus \abbr{L3} short. The dashed line denotes equality; coincident points overlap. All conditions use a 65,536-token ceiling. Both panels use DeepSeek (family A).}
\label{fig:concise}
\end{figure*}

For both DeepSeek settings, \abbr{L3} short requests a one-line complete design, while \abbr{L1} padded requests a longer reason discussing arms and seeds but retains the action-name parser. All conditions share the 65,536-token ceiling; actual response lengths are not matched. We measure how much of the \abbr{L1}-to-\abbr{L3} gain each control recovers, with 4,000 paired state-bootstrap resamples. A complete design can also be concise. Figure~\ref{fig:concise} compares the original and control interfaces state by state. The compact \abbr{L3} interface scores 77.87\% for DeepSeek/low and 75.14\% for DeepSeek/max, with full control results reported in Appendix~\ref{app:controls}. Relative to the \abbr{L1}-to-\abbr{L3} gain in the same setting, it retains 101.5\% at low effort and 101.5\% at maximum effort. Concise parameter specification is therefore sufficient to preserve the observed gain in this benchmark.

By contrast, \abbr{L1} padded scores 13.36\% at low effort and 13.21\% at maximum effort. One empty response accounts for the latter reduction. Because its parser extracts the action name even when the reason mentions design details, elaborating the reason without exposing those details to the executor leaves the scored design interface unchanged. Together, the controls support compact, accessible design fields as an effective interface choice. Their estimand is the performance of the complete request--parser interface at a shared generation ceiling.

\subsubsection{Prediction Field and Reasoning Effort}
Adding the prediction field changes the request from \abbr{L3} to \abbr{L4} while retaining the design scoring rule. On the original 0--1 score scale, the family-aggregated \abbr{L3}-to-\abbr{L4} differences are $-0.0082$ for DeepSeek and $-0.0365$ for Qwen3.8-27B. Neither contrast is significant after Holm adjustment. Similarly, for DeepSeek, low-to-max differences at \abbr{L3} and \abbr{L4} are $-0.0269$ and $-0.0444$, also nonsignificant after adjustment. These comparisons do not establish equivalence between the conditions. The principal design gain is already present at \abbr{L3}, where the executor receives a complete experiment.

\subsection{Discussion}
\label{sec:discussion}
The main comparisons connect explicit design specification with higher affordable information efficiency. Relative to action-name baselines, complete-design interfaces expose the interventions and replication count that determine the evaluated experiment. Concise requests preserve this gain, while the design-format control shows that a complete design can be communicated through different field representations. Beyond simulated efficiency, the constructed dataset adds evidence of agreement with scientific judgments. Its ordering differences also clarify the role of completion because readers evaluate the displayed proposal while the metric includes the declared executor policy. These findings support an evaluation of the information delivered by an interface, with executor sensitivity and deterministic design calibration detailed in Appendix~\ref{app:diagnostics}.

\FloatBarrier
\section{Conclusion}
\label{sec:conclusion}
We formulate action construction as the evaluation of how a research agent turns scientific intent into an executable experiment. Reading sets expose executor completion, and per-design budget gating measures affordable information efficiency under a common cost rule. The accompanying author-constructed research-action evaluation dataset provides traceable source records for 60 scenarios and 240 candidate actions. Across 30 synthetic states, complete-design interfaces increase affordable efficiency by \RelativeGainA\% and \RelativeGainB\% relative to the action-name baseline in the two model families. The constructed dataset provides complementary evidence of positive agreement with a six-model panel and 18 human participants. Together, the framework and datasets provide a reproducible basis for assessing the design information delivered by research agents and the execution choices their instructions leave open.

Future work can connect these measurements to realised hypothesis resolution by executing the proposed experiments and extending the action grammar to data collection and model selection. The current evidence concerns simulated utility and judgments on a constructed dataset whose scenarios can share source papers. Broader source coverage and source-level validation would therefore strengthen transfer assessment. Content-fixed comparisons with a shared semantic extractor could separate representation effects from design content, and assessments of causal explanations could complement the parameter-based score. Prediction and effort comparisons do not establish additional gains under the current conditions, motivating tests of when these fields contribute useful design information. An explicit cost report could further require agents to calculate expected cost under Eq.~\eqref{eq:cost}, check the budget, and revise unaffordable proposals before acceptance.

\clearpage

\input{references}
\clearpage\appendices
\FloatBarrier\section{Reproducibility and Protocol Details}
\label{app:reproduce}

The action interfaces request different amounts of experimental information. We specify their requested fields and executor interpretations in Table~\ref{tab:levels}. These definitions are fixed throughout the main comparison and specify what the parser can pass to the evaluator.

\begin{table*}[!b]
\centering\small
\caption{Action interfaces and their executable interpretation. Fields are requested from the model; scoring follows the fixed parser for each interface.}
\label{tab:levels}
\begin{tabularx}{\textwidth}{l l X X}
\toprule
Level & Interface & Requested fields & Executor interpretation \\
\midrule
\abbr{L1} & Action name & ACTION, REASON & A fixed set of generic designs; design details in REASON are not extracted. \\
\abbr{L2} & Target & TARGET, WHY & Target-conditioned interventions; seed count completed by the executor. \\
\abbr{L3} & Complete design & TARGET, ARMS, SEEDS, SUCCESS & One design when the scored fields parse and satisfy the domain rules. \\
\abbr{L4} & Design + prediction & \abbr{L3} fields plus PREDICTION & The same design scoring rule; the prediction is stored for inspection. \\
\bottomrule
\end{tabularx}
\end{table*}

Provider temperature and seed defaults are retained. The protocol, prompts, research states, and primary scoring rule were fixed before collection. The original interface study includes its 90 preflight responses once. The separate prompting comparison likewise includes its 24 pilot responses once, comprising 18 final designs and six drafts. The interface collection contributes 1,440 main responses, 360 concise/expanded controls, and 180 design-format controls. The separate complete-design comparison contributes 540 final responses and 180 first-turn drafts, giving 2,520 final items and 2,700 responses across the reported analyses. Eight final answers are invalid: four are empty (three with a length termination flag), three contain invalid target strings, and one specifies an out-of-domain design. The parser reads the final answer field without falling back to a reasoning trace; all eight score zero. One DeepSeek first-turn draft in the separate comparison reaches the generation ceiling, and its subsequent final design is retained and scores as a valid response.

The 12 planned paired contrasts comprise \abbr{L1}--\abbr{L2}, \abbr{L1}--\abbr{L3}, \abbr{L1}--\abbr{L4}, and \abbr{L3}--\abbr{L4} for each family, averaging its two settings, plus four DeepSeek low--max comparisons. We use paired $t$-tests with Holm adjustment across the 12 contrasts, differences on the original 0--1 score scale, and paired standardised effects $d_z$. Confidence intervals use 5,000 percentile-bootstrap resamples over states (seed 3). Relative increases divide the difference between family means by the baseline family mean, resampling the paired states for both terms. Concise-control gain ratios use 4,000 paired resamples (seed 7).

All synthetic results share the same states, cost rule, and budget-gated evaluator, while the interface study and the complete-design prompting comparison were collected in separate batches. We retain the complete requests, responses, research states, parser, scorer, and analysis code for reproducibility. Python scripts compute the reported statistics and generate every figure and result table. Numerical consistency checks verify these outputs against the response-level scores. For traceability, each response is linked to its request and evaluation settings. Collection contains 1,530 DeepSeek and 1,170 Qwen3.8-27B responses, including the first turns of self-refinement. At most three logged retries are permitted for transport errors, rate limits, or server failures; responses are not retried because of their quality. Request checks cover the model settings, generation limits, and prompt construction. The analysis retains all 2,520 final scored items, and response-level cryptographic checksums link each score to its archived source.

The scorer enumerates all 693 designs for each of the 30 states and caches their information, cost, and affordability. To verify this computation independently, an information-model audit checks 120 cases using a separate numerical computation; the maximum discrepancy is $2.96\times10^{-8}$. We additionally recompute the original interface and control condition means and confidence intervals, the 12 paired contrasts and Holm corrections, and the concise-control gain ratios from item-level scores. The complete-design prompting results and their six planned contrasts are recomputed separately from the corresponding response-level scores. These checks establish the numerical consistency of the analysis.

The resolver chooses the intervention set for a stated target by combining a target-focused distribution with a uniform component. The target mass is $0.8$ before this mixture; the remaining five hypotheses each receive $0.04$. The component-scan sharpness is $0.75$, and the default selects two interventions with seed alternatives $2$ and $8$. Without a target, the component-scan candidates are $(\{0\},2)$, $(\{0,1\},2)$, $(\{0,1,2,3\},2)$, and $(\{0\},4)$. Ties in the cost-based mode split preserve candidate order. Complete arms and seed fields bypass this completion step.

Table~\ref{tab:quality} distinguishes successful parsing from the stronger condition that \emph{every} admitted reading is feasible. For a complete design, the all-feasible count is the number of parseable outputs within budget. For an incomplete action, however, some readings may exceed the budget even when the mean cost is affordable. The primary score gates each reading individually.

\begin{table}[!htbp]
\centering\footnotesize\setlength{\tabcolsep}{2.9pt}
\caption{Main-matrix response quality. Parsed and all-feasible columns count requests out of 90. The last column averages the fraction of feasible readings over all requests, counting invalid answers as zero. A denotes DeepSeek and B denotes Qwen3.8-27B. The label na means that the reasoning-effort parameter is omitted.}
\label{tab:quality}\input{generated/quality}
\end{table}

Table~\ref{tab:contrasts} reports the 12 planned paired contrasts in the original 0--1 score units, including adjusted significance levels.

\begin{table}[!htbp]
\centering\footnotesize\setlength{\tabcolsep}{2.5pt}
\caption{All 12 planned paired contrasts on the primary score. \abbr{L1}--\abbr{L4} contrasts pool the two tested settings within a family, averaging by state before testing; effort contrasts compare DeepSeek/low with DeepSeek/max. $\Delta Q_B$ uses the original 0--1 score scale. The $p$ values use one 12-test Holm family. A denotes DeepSeek and B denotes Qwen3.8-27B.}
\label{tab:contrasts}\input{generated/contrasts}
\end{table}

\FloatBarrier\section{Complete-Design Prompting Comparisons}
\label{app:complete-baselines}

The prompting comparison uses three task-specific adaptations with the same complete-design fields and scoring rule. Think-then-act makes a single construction decision without a tool-observation loop, while Plan-and-Solve requests an explicit plan before the design. Self-Refine uses two model calls, one for a draft and one for critique and revision, and only the final design enters the comparison. The cost model in Eq.~\eqref{eq:cost} measures the proposed experiment and does not include model-call charges or generation tokens.

The Qwen3.8-27B comparison omits the reasoning-effort parameter and includes all 90 final responses for each method. Planning and self-refinement achieve 68.20\% and 70.05\%, respectively, compared with 60.98\% for \abbr{L3}; think-then-act achieves 55.92\%. Thus, the relative ordering of the complete-design strategies differs across the two model settings, as shown in Table~\ref{tab:prompts-qwen}. One think-then-act design and one self-refined design exceed their state budgets, and another self-refined response has an invalid target string. These three responses remain in the denominator and score zero. All 270 DeepSeek final designs are parseable and budget-feasible, and none of the 540 final responses terminates at the generation ceiling.

\begin{table}[!htbp]
\centering\footnotesize
\caption{Complete-design prompting comparison on Qwen3.8-27B with the reasoning-effort parameter omitted. Efficiency is affordable efficiency share (\%). Every method includes 30 states and three responses per state. Failed outputs remain in the denominator. Bold method names identify our four study interfaces. Bold numerical values mark the highest mean.}
\label{tab:prompts-qwen}\input{generated/prompt_baselines_qwen}
\end{table}

For the six baseline-versus-\abbr{L3} contrasts across both model settings, we average the three repetitions within each state and apply paired $t$-tests with one six-test Holm correction. This testing family is separate from the 12 interface contrasts. For DeepSeek, all three adjusted $p$ values are 0.538. For Qwen3.8-27B, the adjusted $p$ values are 0.324, 0.152, and 0.027 for think-then-act, planning, and self-refinement, respectively. Model identifiers and request settings match across collection batches, but provider changes between batches are not controlled. The comparisons therefore describe the observed efficiency of these instantiated strategies rather than a batch-controlled causal effect of prompting alone.

\FloatBarrier\section{A Fixed Illustrative State}
\label{app:example}
Table~\ref{tab:example} illustrates the scoring procedure using state 0, repetition 0, and DeepSeek/low, selected by index. The observed dataset differences are $+1.5$, $+2.6$, $-0.3$, and $+3.5$, and the budget is $8$. \abbr{L3} specifies interventions 1 and 3 at eight seeds, costing $5.00$. In comparison, \abbr{L4} specifies all six interventions at nine seeds, costing $7.75$. This example illustrates the relation between a response, its resolved design, and its score; the main conclusions use the complete state panel.

\begin{table}[!htb]
\centering\footnotesize\setlength{\tabcolsep}{4pt}
\caption{State 0, repetition 0, DeepSeek/low. Mean cost is averaged over the reading set for \abbr{L1}/\abbr{L2}. Scores apply the same per-reading gate as every other result.}
\label{tab:example}\input{generated/example}
\end{table}

\FloatBarrier\section{Constructed Dataset and External Judgment Details}
\label{app:external}

\begin{table*}[!t]
\centering\small
\caption{External agreement on the scenario-level IG/cost measure. Human ranks are averaged within a scenario before correlation.}
\label{tab:external}\input{generated/external}
\end{table*}

The author-constructed research-action evaluation dataset uses published ablations and component comparisons as source material. Its 60 scenarios refer to 29 distinct source-paper identifiers; each carries a DOI or arXiv identifier and a source location. Reusing a paper allows distinct experimental questions or settings to be represented. Construction uses AI-assisted drafting, source checks, and revision, after which the scenarios are fixed before evaluation. Independent human validation of every scenario is not established. Source-linked observations, proposed intervention recipes, and project assumptions occupy separate fields. The priors, abstract resource budgets, and simulator mechanism mappings are evaluation assumptions rather than estimates fitted to the source experiments; we report no measured follow-up outcomes for the 360 proposed interventions.

Candidate actions are constructed by deterministic rules without additional model calls. Each scenario has three parameterised proposals and one action-name proposal. The parameterised proposals vary target, intervention pairing, and replication, using distinct two-intervention sets; the action-name proposal is resolved by the declared executor. Construction labels are not treated as empirical quality labels. Source identities and private construction labels are omitted from judgment displays, which present the scenario and candidate texts. The source records, construction procedure, and source-to-scenario mapping make these adaptations traceable. Scenario-bootstrap intervals in the reported analysis are conditional on this constructed collection and do not estimate variation across independently sampled source papers.

The evaluation panel contains six open-weight models: Mistral-7B-Instruct-v0.3, GLM-4-9B-chat, Granite-3.3-8B-Instruct, InternLM2.5-7B-chat, Yi-1.5-6B-Chat, and Phi-4-mini-instruct, run in bf16 without quantisation. Each ranks the four actions in each scenario across four cyclic position assignments, yielding 1,440 responses. Every action occupies each displayed letter position once, addressing the option-order sensitivity documented in prior work \cite{ref28}. Decoding is constrained to valid ranking strings while leaving every allowed ordering and tie pattern available. This constrains output format rather than prescribing a preferred action. For each model, action ranks are averaged over the four positions; the panel rank then averages the six model ranks. We report pooled correlations across all 240 actions and mean within-scenario correlations across each set of four actions. A constant rank vector has undefined Spearman correlation. Yi has one constant position-averaged rank vector, so its mean within-scenario correlation uses the remaining 59 scenarios. The pooled and panel results reported in Table~\ref{tab:panel-models} retain all of its responses.

Scenario order, pair order, and left--right assignment vary by participant. Accordingly, participant-specific maps decode the choices before aggregation. Within each scenario, participant ranks are averaged and correlated with the metric, and the resulting correlations are averaged across the 60 scenarios.

The external judgments were collected separately from the synthetic benchmark. The panel decoding grammar admits 75 weak orderings with ties and 192 accepted textual spellings. These are output-format constraints, and the analysis averages positions and clusters by scenario.

Human records are marked as participant-entered and human-self-reported. All 18 sessions were completed sequentially on one device. Under the recorded quality rules, no participant or participant--scenario cell was excluded; all 864 judgments remain. Twelve sessions finished in under ten minutes, which the collection protocol treated as a review flag rather than an automatic exclusion. The reviewed collection was retained. These records establish the reported response counts and calculations; participant research experience is self-reported.

The 144 participant--scenario cells are aggregated into scenario ranks before correlation. The panel--human interval uses 5,000 scenario-bootstrap resamples of the stored ranks (seed 3). Individual pairwise comparisons are not treated as independent bootstrap units. Table~\ref{tab:external} reports the three aggregate correlations with their distinct aggregation units.

\FloatBarrier\section{Detailed Interface Controls}
\label{app:controls}
The concise and expanded controls use the same generation ceiling as the main interfaces. We report their scores and gain-retention estimates in Table~\ref{tab:controls}. The compact design retains the main construction gain, while an expanded reason leaves the action-name parser unchanged.

\begin{table}[!htbp]
\centering\footnotesize\setlength{\tabcolsep}{2.4pt}
\caption{Concise/expanded interface controls. Scores are relative to the affordable optimum (\%). Gain retained is $(Q_{\rm short}-Q_{\rm L1})/(Q_{\rm L3}-Q_{\rm L1})$. A denotes DeepSeek.}
\label{tab:controls}\input{generated/controls}
\end{table}

The design-format control requests target, interventions, and repetition count in JavaScript Object Notation (JSON), with 90 final outputs across 30 states per setting. It scores 68.0\% on DeepSeek/max and 63.9\% on Qwen3.8-27B/na under the same budget-gated score. Compared with the complete-design scores reported in Table~\ref{tab:main}, it ranks lower on DeepSeek/max and higher on Qwen3.8-27B/na. Because the requests and field representations differ, this comparison does not isolate a causal effect of JSON formatting.

\FloatBarrier\section{Executor Sensitivity and Design Diagnostics}
\label{app:diagnostics}

\subsection{Information-Gap Definition}

\label{sec:diagnostic}
An auxiliary decomposition locates information loss on the IG scale. Let $I^*$ be the largest IG among affordable designs and $I_t^*$ the largest IG among affordable designs containing the intervention associated with target $t$. For an unstated target, the implementation uses the highest-prior mechanism. With $\overline{I}(a)$ the mean, ungated IG across readings, we define the target-restriction gap $G_{\mathrm{tgt}}$ and resolution gap $G_{\mathrm{res}}$:
\begin{align}
G_{\mathrm{tgt}}&=I^*-I_t^*,\\
G_{\mathrm{res}}&=I_t^*-\overline{I}(a),\\
G_{\mathrm{total}}&=G_{\mathrm{tgt}}+G_{\mathrm{res}}=I^*-\overline{I}(a).
\end{align}
The target-restriction gap measures the reduction in optimal affordable IG imposed by the target requirement; the resolution gap compares this restricted optimum with the mean IG of the admitted readings. These are information-domain diagnostics, distinct from the cost-normalised score in Eq.~\eqref{eq:score}. The resolution term can be negative when an admitted reading exceeds the budget or falls outside the target-restricted domain. We therefore retain signed values and report their denominators. Table~\ref{tab:diagnostics} reports these diagnostics for all main conditions.

\begin{table}[!htbp]
\centering\small\setlength{\tabcolsep}{3pt}
\caption{Signed information-gap diagnostics (nats), averaged over the $N$ parsed outputs in each main condition. The final column counts negative resolution terms below $-10^{-10}$. These quantities decompose ungated IG, not the primary affordable-efficiency score. A denotes DeepSeek and B denotes Qwen3.8-27B. The label na means that the reasoning-effort parameter is omitted.}
\label{tab:diagnostics}\input{generated/diagnostics}
\end{table}

\subsection{Completion Policies and Design Calibration}

We next test whether a score depends on executor completion and calibrate complete specification against design quality. The recorded actions are rescored under conservative, default, and permissive completion policies. Deterministic design policies provide references without additional model calls. Fully specified \abbr{L3}/\abbr{L4} designs have singleton reading sets, so their scores remain identical under conservative, default, and permissive executor modes, including the zero scores of invalid answers. Table~\ref{tab:sensitivity} shows how \abbr{L1}/\abbr{L2} scores change when the executor favours cheaper or more expensive completions. Explicit design fields therefore make the scored experiment stable across the tested completion policies.

\begin{table}[!htbp]
\centering\small\setlength{\tabcolsep}{3pt}
\caption{Executor sensitivity: affordable efficiency share (\%) with invalid answers counted as zero in every mode. A denotes DeepSeek and B denotes Qwen3.8-27B. The label na means that the reasoning-effort parameter is omitted.}
\label{tab:sensitivity}\input{generated/sensitivity}
\end{table}

We compare exhaustive affordable search over the full domain, exhaustive search requiring the highest-prior target, a uniform distribution over affordable designs, a single-arm policy on the highest-prior target, and a fixed six-arm policy. Each uses Eq.~\eqref{eq:score}. Table~\ref{tab:deterministic} places these policies and model interfaces on a common design scale. A uniform distribution over affordable designs reaches 44.3\%, a top-prior single-arm/two-seed design 10.7\%, and the fixed six-arm/five-seed policy 15.2\%. The full-domain oracle reaches 100\% by definition. Restricting the exhaustive search to designs containing the top-prior target also reaches 100\% on these states. These policies demonstrate the benchmark's optimisation headroom and distinguish complete specification from optimal design selection. The measured contribution is the interface's ability to deliver an explicit, valuable experiment under the budget.

\begin{table}[!htbp]
\centering\footnotesize\setlength{\tabcolsep}{3pt}
\caption{Deterministic design calibration under the primary budget-gated score. All policies use the same 30 states and the full 693-design catalogue. Uniform performance is an exact average over affordable designs.}
\label{tab:deterministic}\input{generated/deterministic}
\end{table}

\subsection{Detailed Sensitivity and Gap Results}
The recomputed scores in Table~\ref{tab:sensitivity} use the recorded reading sets under all three executor modes. Complete designs remain invariant because each admits one reading. For incomplete actions, mode changes combine a different design cost with a different chance of falling within the budget. Mode names refer to the cost ordering, not a guaranteed score ordering.

The information-gap results in Table~\ref{tab:diagnostics} use successfully parsed outputs. This diagnostic subset is explicit; all primary scores and tests continue to include failures. Negative resolution terms are retained. The target condition requires the chosen target's intervention to be present, while permitting other interventions; it can leave the globally best affordable design accessible. This explains the small target-restriction gaps in the full domain and makes their interpretation conditional on the stated target restriction.

\end{document}

%% file: generated/numbers.tex
\newcommand{\DesignFeasible}{715}
\newcommand{\DesignTotal}{720}
\newcommand{\FeasiblePct}{99.31}

\newcommand{\ScoreLone}{13.4}
\newcommand{\ScoreAlowLthree}{76.9}
\newcommand{\ScoreAmaxLthree}{74.2}
\newcommand{\ScoreBnaLthree}{61.0}
\newcommand{\ScoreBlowLthree}{53.3}

\newcommand{\RhoS}{0.566}
\newcommand{\RhoHuman}{0.733}
\newcommand{\RhoSH}{0.824}
\newcommand{\RelativeGainA}{465.98}

\newcommand{\FamilyBaseA}{13.36}
\newcommand{\FamilyCompleteA}{75.59}
\newcommand{\RelativeGainB}{327.91}

\newcommand{\FamilyBaseB}{13.36}
\newcommand{\FamilyCompleteB}{57.15}

%% file: generated/main.tex
\begin{tabular}{lrrrr}
\toprule
Setting & \shortstack{\abbr{L1}\\Action-name\\interface} & \shortstack{\abbr{L2}\\Target\\interface} & \shortstack{\abbr{L3}\\Complete-design\\interface} & \shortstack{\abbr{L4}\\Design-with-prediction\\interface} \\
\midrule
DeepSeek (A)/low & 13.4 & 25.4 & 76.9 & 77.0 \\
DeepSeek (A)/max & 13.4 & 25.3 & 74.2 & 72.6 \\
Qwen3.8-27B (B)/na & 13.4 & 25.2 & 61.0 & 57.7 \\
Qwen3.8-27B (B)/low & 13.4 & 25.0 & 53.3 & 49.3 \\
\bottomrule
\end{tabular}

%% file: generated/prompt_baselines.tex
\begin{tabular*}{\columnwidth}{@{\extracolsep{\fill}}lr@{}}
\toprule
\multicolumn{2}{c}{\textbf{DeepSeek (reasoning effort = max)}} \\
\midrule
Method & Efficiency (\%) $\uparrow$ \\
\midrule
Think then act (ReAct-inspired)~\cite{yao2023react} & 70.97 \\
Plan-and-Solve adaptation~\cite{wang2023plan} & 70.98 \\
One-round Self-Refine~\cite{madaan2023selfrefine} & 71.88 \\
\addlinespace[3pt]
\textbf{Action-name interface (L1)} & 13.36 \\
\textbf{Target interface (L2)} & 25.29 \\
\textbf{Complete-design interface (L3)} & \textbf{74.25} \\
\textbf{Design-with-prediction interface (L4)} & 72.56 \\
\bottomrule
\end{tabular*}

%% file: generated/panel_models.tex
\begin{tabular}{lrr}
\toprule
Model & Pooled $\rho$ & Mean within-scenario $\rho$ \\
\midrule
Yi-1.5-6B-Chat & 0.309 & 0.258 \\
InternLM2.5-7B-chat & 0.588 & 0.594 \\
GLM-4-9B-chat & 0.570 & 0.568 \\
Granite-3.3-8B-Instruct & 0.502 & 0.467 \\
Phi-4-mini-instruct & 0.244 & 0.217 \\
Mistral-7B-Instruct-v0.3 & 0.262 & 0.252 \\
\midrule
Model panel (mean rank) & 0.566 & 0.545 \\
\bottomrule
\end{tabular}

%% file: generated/quality.tex
\begin{tabular}{llrrr}
\toprule
Setting & Level & Parsed /90 & All feasible /90 & Feasible readings (\%) \\
\midrule
A/low & \abbr{L1} & 90 & 57 & 90.8 \\
A/low & \abbr{L2} & 90 & 57 & 81.7 \\
A/low & \abbr{L3} & 90 & 90 & 100.0 \\
A/low & \abbr{L4} & 90 & 90 & 100.0 \\
A/max & \abbr{L1} & 90 & 57 & 90.8 \\
A/max & \abbr{L2} & 90 & 57 & 81.7 \\
A/max & \abbr{L3} & 90 & 90 & 100.0 \\
A/max & \abbr{L4} & 90 & 90 & 100.0 \\
B/na & \abbr{L1} & 90 & 57 & 90.8 \\
B/na & \abbr{L2} & 90 & 57 & 81.7 \\
B/na & \abbr{L3} & 88 & 88 & 97.8 \\
B/na & \abbr{L4} & 90 & 88 & 97.8 \\
B/low & \abbr{L1} & 90 & 57 & 90.8 \\
B/low & \abbr{L2} & 89 & 57 & 81.1 \\
B/low & \abbr{L3} & 90 & 90 & 100.0 \\
B/low & \abbr{L4} & 90 & 89 & 98.9 \\
\bottomrule
\end{tabular}

%% file: generated/contrasts.tex
\begin{tabular}{llrrr}
\toprule
Family & Contrast & $\Delta Q_B$ & $d_z$ & $p_{\mathrm{Holm}}$ \\
\midrule
A & \abbr{L1} $\to$ \abbr{L2} & 0.1201 & 0.76 & 0.00197 \\
A & \abbr{L1} $\to$ \abbr{L3} & 0.6224 & 3.47 & 7.42e-17 \\
A & \abbr{L1} $\to$ \abbr{L4} & 0.6142 & 3.49 & 6.44e-17 \\
A & \abbr{L3} $\to$ \abbr{L4} & -0.0082 & -0.13 & 1 \\
B & \abbr{L1} $\to$ \abbr{L2} & 0.1176 & 0.74 & 0.00245 \\
B & \abbr{L1} $\to$ \abbr{L3} & 0.4380 & 2.89 & 7.66e-15 \\
B & \abbr{L1} $\to$ \abbr{L4} & 0.4015 & 3.21 & 5.03e-16 \\
B & \abbr{L3} $\to$ \abbr{L4} & -0.0365 & -0.29 & 0.723 \\
A & low $\to$ max, \abbr{L1} & 0.0000 & -- & 1 \\
A & low $\to$ max, \abbr{L2} & -0.0016 & -0.21 & 1 \\
A & low $\to$ max, \abbr{L3} & -0.0269 & -0.17 & 1 \\
A & low $\to$ max, \abbr{L4} & -0.0444 & -0.29 & 0.723 \\
\bottomrule
\end{tabular}

%% file: generated/prompt_baselines_qwen.tex
\begin{tabular*}{\columnwidth}{@{\extracolsep{\fill}}lr@{}}
\toprule
Method & Efficiency (\%) $\uparrow$ \\
\midrule
Think then act (ReAct-inspired)~\cite{yao2023react} & 55.92 \\
Plan-and-Solve adaptation~\cite{wang2023plan} & 68.20 \\
One-round Self-Refine~\cite{madaan2023selfrefine} & \textbf{70.05} \\
\addlinespace[3pt]
\textbf{Action-name interface (L1)} & 13.36 \\
\textbf{Target interface (L2)} & 25.23 \\
\textbf{Complete-design interface (L3)} & 60.98 \\
\textbf{Design-with-prediction interface (L4)} & 57.70 \\
\bottomrule
\end{tabular*}

%% file: generated/example.tex
\begin{tabular}{llrrr}
\toprule
Level & Arms & Seeds & Cost & Score (\%) \\
\midrule
\abbr{L1} & Reading set & -- & 3.08 & 8.40 \\
\abbr{L2} & Reading set & -- & 3.95 & 27.66 \\
\abbr{L3} & 1,3 & 8 & 5.00 & 55.51 \\
\abbr{L4} & 0,1,2,3,4,5 & 9 & 7.75 & 94.34 \\
\bottomrule
\end{tabular}

%% file: generated/external.tex
\begin{tabular}{llr}
\toprule
Comparison & Aggregation & Spearman $\rho$ \\
\midrule
Model panel / metric & Pooled across 240 actions & 0.566 \\
Humans / metric & Mean within 60 scenarios & 0.733 \\
Model panel / humans & Mean within 60 scenarios & 0.824 \\
\bottomrule
\end{tabular}

%% file: generated/controls.tex
\begin{tabular}{lrrrrr}
\toprule
A effort & \abbr{L1} & \abbr{L1} padded & \abbr{L3} short & \abbr{L3} full & Gain retained (\%) \\
\midrule
low & 13.36 & 13.36 & 77.87 & 76.94 & 101.5 \\
max & 13.36 & 13.21 & 75.14 & 74.25 & 101.5 \\
\bottomrule
\end{tabular}

%% file: generated/diagnostics.tex
\begin{tabular}{llrrrrr}
\toprule
Setting & Level & $N$ & $G_{\rm tgt}$ & $G_{\rm res}$ & $G_{\rm total}$ & $G_{\rm res}<0$ \\
\midrule
A/low & \abbr{L1} & 90 & 0.0000 & 0.5249 & 0.5249 & 0 \\
A/low & \abbr{L2} & 90 & 0.0000 & 0.3877 & 0.3877 & 14 \\
A/low & \abbr{L3} & 90 & 0.0000 & 0.1257 & 0.1257 & 0 \\
A/low & \abbr{L4} & 90 & 0.0000 & 0.1344 & 0.1344 & 0 \\
A/max & \abbr{L1} & 90 & 0.0000 & 0.5249 & 0.5249 & 0 \\
A/max & \abbr{L2} & 90 & 0.0000 & 0.3887 & 0.3887 & 13 \\
A/max & \abbr{L3} & 90 & 0.0000 & 0.1310 & 0.1310 & 0 \\
A/max & \abbr{L4} & 90 & 0.0000 & 0.1160 & 0.1160 & 0 \\
B/na & \abbr{L1} & 90 & 0.0000 & 0.5249 & 0.5249 & 0 \\
B/na & \abbr{L2} & 90 & 0.0006 & 0.3887 & 0.3894 & 12 \\
B/na & \abbr{L3} & 88 & 0.0000 & 0.2889 & 0.2889 & 0 \\
B/na & \abbr{L4} & 90 & 0.0000 & 0.3027 & 0.3027 & 2 \\
B/low & \abbr{L1} & 90 & 0.0000 & 0.5249 & 0.5249 & 0 \\
B/low & \abbr{L2} & 89 & 0.0000 & 0.3951 & 0.3951 & 7 \\
B/low & \abbr{L3} & 90 & 0.0028 & 0.3486 & 0.3514 & 0 \\
B/low & \abbr{L4} & 90 & 0.0006 & 0.3842 & 0.3848 & 1 \\
\bottomrule
\end{tabular}

%% file: generated/sensitivity.tex
\begin{tabular}{llrrr}
\toprule
Setting & Level & Conservative & Default & Permissive \\
\midrule
A/low & \abbr{L1} & 10.81 & 13.36 & 15.90 \\
A/low & \abbr{L2} & 14.37 & 25.45 & 36.52 \\
A/low & \abbr{L3} & 76.94 & 76.94 & 76.94 \\
A/low & \abbr{L4} & 76.99 & 76.99 & 76.99 \\
A/max & \abbr{L1} & 10.81 & 13.36 & 15.90 \\
A/max & \abbr{L2} & 14.28 & 25.29 & 36.30 \\
A/max & \abbr{L3} & 74.25 & 74.25 & 74.25 \\
A/max & \abbr{L4} & 72.56 & 72.56 & 72.56 \\
B/na & \abbr{L1} & 10.81 & 13.36 & 15.90 \\
B/na & \abbr{L2} & 14.18 & 25.23 & 36.27 \\
B/na & \abbr{L3} & 60.98 & 60.98 & 60.98 \\
B/na & \abbr{L4} & 57.70 & 57.70 & 57.70 \\
B/low & \abbr{L1} & 10.81 & 13.36 & 15.90 \\
B/low & \abbr{L2} & 13.71 & 25.01 & 36.32 \\
B/low & \abbr{L3} & 53.32 & 53.32 & 53.32 \\
B/low & \abbr{L4} & 49.30 & 49.30 & 49.30 \\
\bottomrule
\end{tabular}

%% file: generated/deterministic.tex
\begin{tabular}{lr}
\toprule
Calibration policy & Score (\%) \\
\midrule
Full-domain affordable oracle & 100.0 \\
Top-prior target + exhaustive design search & 100.0 \\
Uniform over affordable designs & 44.3 \\
Top-prior single arm, two seeds & 10.7 \\
Fixed six arms, five seeds & 15.2 \\
\bottomrule
\end{tabular}